\UseRawInputEncoding
\def\year{2022}\relax
\documentclass[letterpaper]{article} 
\usepackage{aaai22}  
\usepackage{times}  
\usepackage{helvet}  
\usepackage{courier}  
\usepackage[hyphens]{url}  
\usepackage{graphicx} 
\usepackage{natbib}  
\usepackage{caption} 
\DeclareCaptionStyle{ruled}{labelfont=normalfont,labelsep=colon,strut=off} 
\usepackage{algorithm}
\usepackage{algorithmic}
\usepackage{hyperref}
\hypersetup{
    colorlinks=true,
    linkcolor=black,
    filecolor=black,
    urlcolor=blue,
    citecolor=black,
}

\usepackage{newfloat}
\usepackage{listings}
\usepackage{booktabs}
\floatstyle{ruled}
\newfloat{listing}{tb}{lst}{}
\floatname{listing}{Listing}
\usepackage{bbding}
\usepackage{amsfonts}
\usepackage{eucal}
\usepackage{multirow}
\usepackage{amsmath}
\usepackage{color}
\usepackage{gensymb}
\usepackage{threeparttable}
\usepackage[switch]{lineno}
\definecolor{brown}{RGB}{96,56,17}

\title{Contrastive Spatio-Temporal Pretext Learning for \\
Self-supervised Video Representation}
\author {
    Yujia Zhang\textsuperscript{\rm 1}\thanks{Corresponding author.},
    Lai-Man Po\textsuperscript{\rm 1},
    Xuyuan Xu\textsuperscript{\rm 2},
    Mengyang Liu\textsuperscript{\rm 2}, \\
    Yexin Wang\textsuperscript{\rm 2},
    Weifeng Ou\textsuperscript{\rm 1},
    Yuzhi Zhao\textsuperscript{\rm 1},
    Wing-Yin Yu\textsuperscript{\rm 1}
}
\affiliations {
    \textsuperscript{\rm 1}Department of Electrical Engineering, City University of Hong Kong, Hong Kong, China\\
    \textsuperscript{\rm 2}AI Technology Center, OVB, Tencent, Shenzhen, China\\
    \{yzhang2383-c,weifengou2-c,yzzhao2-c,wingyinyu8-c\} \\
    @my.cityu.edu.hk, eelmpo@cityu.edu.hk, \{xuyuanxu2,yexin.w\}@gmail.com, myleonliu@tencent.com
}

\usepackage{bibentry}

\begin{document}
\maketitle

\begin{abstract}
    Spatio-temporal representation learning is critical for video self-supervised representation. Recent approaches mainly use contrastive learning and pretext tasks. However, these approaches learn representation by discriminating sampled instances via feature similarity in the latent space while ignoring the intermediate state of the learned representations, which limits the overall performance. In this work, taking into account the degree of similarity of sampled instances as the intermediate state, we propose a novel pretext task - spatio-temporal overlap rate (STOR) prediction. It stems from the observation that humans are capable of discriminating the overlap rates of videos in space and time. This task encourages the model to discriminate the STOR of two generated samples to learn the representations. Moreover, we employ a joint optimization combining pretext tasks with contrastive learning to further enhance the spatio-temporal representation learning. We also study the mutual influence of each component in the proposed scheme. Extensive experiments demonstrate that our proposed STOR task can favor both contrastive learning and pretext tasks. The joint optimization scheme can significantly improve the spatio-temporal representation in video understanding. The code is available at \url{https://github.com/Katou2/CSTP}.
\end{abstract}

\section{Introduction}
\noindent Convolutional Neural Networks (CNNs) have been proven to be successful in supervised video representation learning with numerous human-annotated labels \cite{carreira2017quo, feichtenhofer2019slowfast}. Videos contain more complex spatio-temporal contents and a larger data volume. Billions of unlabeled videos emerge on the Internet every day, making supervised video analysis expensive and time-consuming. Thus, how to effectively learn video representations without annotations is an important yet challenging task. Among effective unsupervised learning methods, self-supervised learning has proven to be a promising methodology \cite{chen2020simple, he2020momentum, feichtenhofer2021large}.

Early video self-supervised learning approaches proposed proper tasks with automatically generated labels, thereby encouraging CNNs to learn the transferable features for downstream tasks without human-annotated labels \cite{fernando2017self, benaim2020speednet, behrmann2021unsupervised}. Recently, with the success of contrastive learning in self-supervised image classification, this method has been widely expended in video self-supervised learning \cite{chen2021rspnet, alwassel2019self, sermanet2018time}.

However, there are obvious limitations in these works. Firstly, previous works only explored discriminating similar features from dissimilar ones while ignoring the intermediate state of learned representations such as the similarity degree, which limits the overall performance. Secondly, less effort has been put on the mutual influence of multiple pretext tasks and various contrastive learning schemes for spatio-temporal representation learning.

\begin{figure}[t]
\centering
\includegraphics[width=8.5cm]{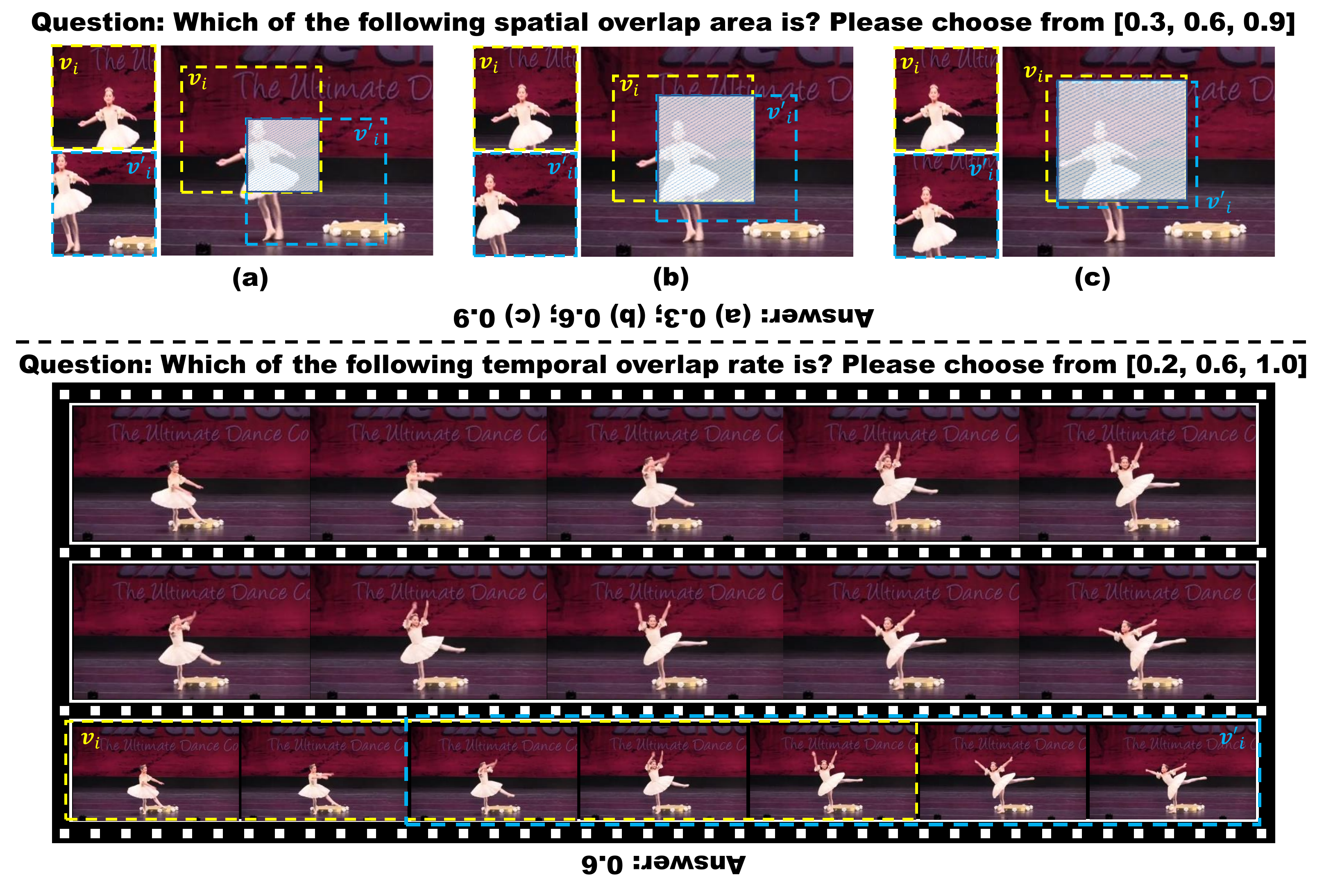}
\caption{Illustration of the proposed overlap rate prediction in space and time. Given a video sample, the first clip $v_i$ (\textcolor{yellow}{Yellow}) is randomly cropped in space and time, the second clip ${v'}_i$ (\textcolor{blue}{Blue}) is randomly cropped according to the size of bounding boxes and time duration of $v_i$ at a random spatial and temporal overlap rate.}
\label{fig1}
\end{figure}

To address these problems, we propose a novel pretext task, i.e., Spatio-Temporal Overlap Rate (STOR) prediction to percept the similarity degree as the intermediate state to favor contrastive learning and propose a joint optimization framework of contrastive learning and multiple pretext tasks to further enhance the spatio-temporal representation learning. It is observed that given a set of two clips at a specific overlap rate, humans can discriminate the overlap rate when providing candidates (see Figure \ref{fig1}). We assume that humans can make it due to their favorable spatio-temporal representation ability. Built upon the observation, we believe that CNNs can learn video representations better by discriminating such overlap rates. The assumption is that CNNs can only succeed in such a spatio-temporal overlap rate reasoning task when it learns representative spatio-temporal features. To the best of our knowledge, this is the first work that attempts to capture the spatio-temporal degree of similarity between generated samples for self-supervised learning. Moreover, we propose a new and effective data augmentation method for the pretext task. This data augmentation method can generate samples with random overlapped spatio-temporal regions, while keeping the randomness of samples.

In order to study the mutual influence of contrastive learning and pretext tasks for better spatio-temporal representation learning, we comprehensively study the joint optimization scheme. Specifically, we study four popular contrastive learning frameworks, i.e., SimCLR \cite{chen2020simple}, MoCo \cite{he2020momentum}, BYOL \cite{grill2020bootstrap} and SimSiam \cite{chen2021exploring} in our joint optimization scheme. In terms of pretext tasks, playback rate prediction has been proven to be successful in video self-supervised learning \cite{wang2020self, jenni2020video}, but it tends to focus on motion pattern thus may not learn spatial pattern well \cite{chen2021rspnet}. To overcome this problem, we combine rotation prediction task to further strengthen spatial appearance features. Finally, a joint optimization framework of STOR, contrastive learning, playback rate prediction and rotation prediction is proposed in this work, namely contrastive spatio-temporal pretext learning (CSTP).

Extensive experimental evaluations on two downstream video understanding tasks demonstrate the effectiveness of the proposed approach. Specifically, several architectures including C3D \cite{tran2015learning}, R(2+1)D \cite{tran2018closer} and S3D \cite{xie2018rethinking} are presented and different weights of the joint learning are explored in this work. The experimental results verifies that the proposed STOR can well cooperate with different contrastive learning frameworks and other pretext tasks. The proposed joint learning framework CSTP outperforms state-of-the-art approaches in the two downstream video understanding tasks.

The main contributions of this work can be summarized as follows:

\begin{itemize}
\item Taking the degree of similarity of training samples into account, we propose a novel pretext task, i.e., spatio-temporal overlap rate prediction for video self-supervised learning. The pretext task can enhance spatio-temporal representation learning via discriminating the overlap regions of the training samples.
\item We propose a joint optimization framework which combines contrastive learning with spatio-temporal pretext tasks. And we conduct comprehensive experiments to study the mutual influence of each component of the framework.
\item Our method achieves state-of-the-art performance on two downstream tasks, action recognition and video retrieval, across two datasets, UCF-101 and HMDB-51. Ablation studies demonstrate the efficacy of the proposed STOR and the mutual influence of contrastive learning and pretext tasks.
\end{itemize}

\section{Related Works}
\subsection{Self-supervised Learning in Images}
Current image self-supervised methods can be briefly grouped into two types of paradigms, i.e., contrastive learning and pretext tasks. Early work \cite{dosovitskiy2015discriminative} studied self-supervised learning by setting each sample in the dataset as a category to train a classification model. However, the approach will be infeasible when the size of dataset is huge. To settle this problem, Wu et al. \cite{wu2018unsupervised} employed a memory bank of previous features of samples to replace the classifier. He et al. \cite{he2020momentum} proposed a dynamic dictionary with a queue for memory bank and a momentum encoder to maintain the consistency of samples. Chen et al. \cite{chen2020simple} used a larger batch size to fully replace the memory bank. Grill et al. \cite{grill2020bootstrap} proposed to use a MLP as feature predictor to replace the memory bank. Recently, Chen and He \cite{chen2021exploring} showed that using stop-gradient method, a Siamese architecture without memory bank and a MLP feature predictor can achieve state-of-the-art results.
Also, self-supervised learning approaches based on pretext tasks have been extensively studied. Examples include recovering the input with appearance transformations, e.g., image colorization \cite{zhao2020scgan, su2020instance}, denoising \cite{vincent2008extracting}. Besides, Pseudo label-based approaches include patch ordering \cite{doersch2015unsupervised}, rotation angle prediction \cite{gidaris2018unsupervised}, frame tracking \cite{wang2015unsupervised}, solving the jigsaw puzzle \cite{noroozi2016unsupervised}.

\subsection{Self-supervised Video Representation Learning}
Researches in video self-supervised learning follow a similar trajectory as image self-supervised learning, which also have two groups of approaches, i.e., pretext tasks and contrastive learning. Video pretext tasks explored natural video properties or statistics as supervision signal on unlabeled data, e.g., frame prediction \cite{vondrick2016anticipating, behrmann2021unsupervised, luo2017unsupervised}, spatio-temporal puzzling \cite{kim2019self}, video statistics \cite{wang2021self}, temporal ordering \cite{misra2016shuffle, yao2021seco}, video playback rate prediction \cite{benaim2020speednet, jenni2020video}, temporal consistency \cite{wang2019learning, jabri2020space}. Recently, inspired by the success of contrastive learning on static images, contrastive learning was expanded in video self-supervised learning \cite{chen2021rspnet, alwassel2019self, sermanet2018time, liu2021temporal}.

Despite the success of contrastive learning and playback rate prediction, contrastive learning approaches just focus on discriminating instances by the similarity of features but ignore the intermediate state of learned representation such as the similarity degree of features, which limits the overall performance. Besides, although pretext tasks and contrastive learning were proven to be successful in video self-supervised learning, few efforts have been placed on the mutual influence of multiple contrastive learning schemes and pretext tasks. Also, there is no guidance for a joint optimization. To settle the first problem, we proposed a novel pretext task, i.e., spatio-temporal overlap rate prediction, to encourage a model to learn the degree of similarity so as to enhance spatio-temporal feature leanring. For the second problem, based on the contrastive learning, we proposed a joint learning with rotation angle prediction to further enhance the spatial representations. Besides, we explore the mutual influence of contrastive learning and provide guidance for the joint optimization.

\section{Methodology}
\subsection{Preliminary-Contrastive Learning Frameworks}
In this subsection, we briefly review four relative state-of-the-art contrastive learning frameworks in this paper. Let $X=\{\mathbf{x_i}\}, i\in[1, N]$ denotes the video training dataset, where $N$ is the total number of videos in the dataset.

\subsubsection{SimCLR \cite{chen2020simple}}\label{ctr}
SimCLR consists of four major steps:

\textbf{Data Augmentation.} A stochastic spatio-temporal data augmentation $\mathcal{T}$ is performed on each input $\left\{\mathbf{x}_i\right\},\ i\in{[1,N}_B]$ in a minibatch, $N_B$ is the minibatch size. Two correlated views $\mathbf{v}_i,\ \mathbf{v}'_i$ are generated from $\mathbf{x}_i$ via by $t\sim\mathcal{T},\ t'\sim\mathcal{T}$. The generated pair $(\mathbf{v}_i,\mathbf{v}'_i)$ is considered as a positive pair. Generated pairs $(\mathbf{v}_i,\mathbf{v}_j)$, $(\mathbf{v}_i,\mathbf{v}'_j)$ are considered as negative pairs, where $i\in{[1,N}_B],j\neq i$.

\textbf{Feature Encoding.} A spatio-temporal CNN model $f_\theta$ that extracts representation vectors from augmented data examples $\left\{\mathbf{v}_i,{\mathbf{v}^\prime}_i\right\},\ i\in{[1,N}_B]$, where $\theta$ is the parameter of the model.

\textbf{Latent Space Projection.}  A multi-layer perception (MLP) projection head $g_\theta$ that maps the representation $\mathbf{h}_i$ to a latent space $\mathbf{z}_i$ which is used for contrastive loss calculation.

\textbf{Contrastive Loss.} With the obtained projected features ${{z}_i},\ i\in{[1,N}_B]$, InfoNCE \cite{oord2018representation} is adopted as the contrastive objective:
\begin{align}
    \mathcal{L}_{ctr}=-\sum_{i=1}^{N_B}\log{\frac{\exp{\left(sim\left(\mathbf{v}_i,{\mathbf{v}'}_i\right)/\alpha\right)}}{\sum_{k\in\left\{\mathbf{v}_j,{\mathbf{v}'}_j\right\},j\neq i}\exp{\left(sim\left(\mathbf{v}_i,k\right)/\alpha\right)}}},
\end{align}
where $sim\left(\mathbf{x},\mathbf{y}\right)=\frac{\mathbf{x}^T\mathbf{y}}{||\mathbf{x}||||\mathbf{y}||}$ that is the cosine similarity calculation of two input vectors.
\subsubsection{MoCo \cite{he2020momentum}}
The architecture of MoCo is similar to SimCLR. The difference between MoCo and SimCLR is that, MoCo builds a queue to store negative samples while SimCLR regards the embeddings in the mini-batch as negative samples. Besides, MoCo uses an explicit momentum encoder which adopts a moving average manner with a momentum parameter $m$.
\begin{align}
    \xi\gets m\xi+(1-m)\theta,
\end{align}
where $\theta$ is the parameter in encoder $f_\theta$, $\xi$ is the parameter in momentum encoder $f_\xi$.

\subsubsection{BYOL \cite{grill2020bootstrap}}
BYOL is a typical contrastive learning method that does not use negative samples in the contrastive loss. BYOL can be seen as a form of MoCo without negative samples but uses an extra MLP predictor $p_\theta$ after the latent space $\mathbf{z}_i$ for predicting latent space ${z'}_i$ to learn the data representation. The objective of BYOL is to minimize the distance between predicted latent features $q_i$ from normal encoder and the projected latent feature ${z'}_i$ of momentum encoder, which is:
\begin{align}
    \mathcal{L}_{ctr}=-\sum_{i=1}^{N_B}{sim(\mathbf{v}_i,{\mathbf{v}'}_i)},
\end{align}

\subsubsection{SimSiam \cite{chen2021exploring}}
SimSiam can be seen as a form of BYOL that does not use a momentum encoder. The difference is that SimSiam uses a Siamese encoder to replace the momentum encoder, but only backpropagate one of the two encoders, i.e., stopping the gradients of the second encoder.

\subsection{Spatio-temporal Overlap Rate Prediction}
\begin{figure}[t]
\centering
\includegraphics[width=8.5cm]{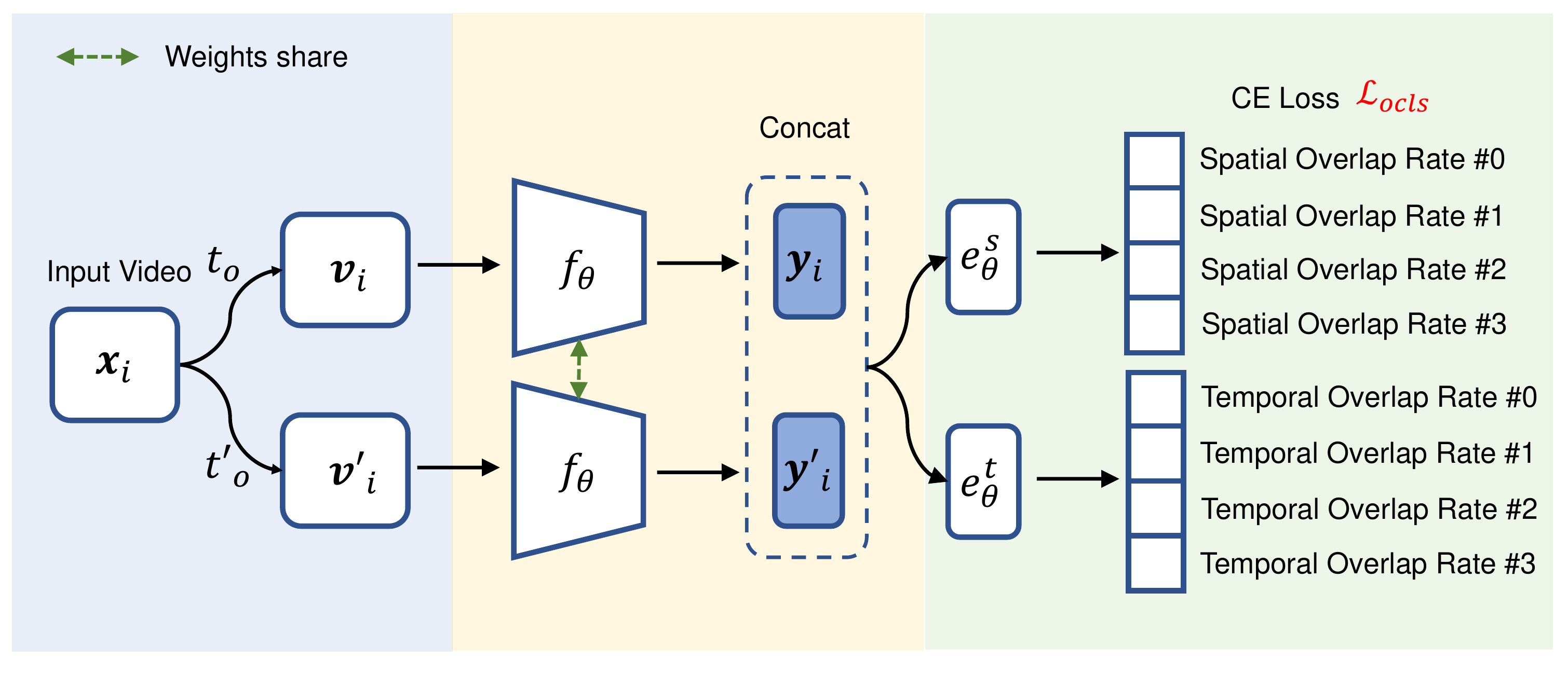}
\caption{An overview of spatio-temporal overlap rate (STOR) prediction.}
\label{fig2}
\end{figure}

Since contrastive learning approaches and existing pretext tasks do not percept the degree of similarity of generated samples while humans can make it (Figure \ref{fig1}). Following the nature of humans, we propose a novel pretext task spatio-temporal overlap rate (STOR) prediction to help models understand the similarity of degree to further enhance the spatio-temporal representation. Following the settings of pretext tasks, our proposed approach encourages CNN model to discriminate different spatio-temporal overlap rates of two training video clips to learn the video representations. The hypothesis is that the network can only succeed in such a spatio-temporal overlap reasoning task when it understands the underlying video content and learns representative spatio-temporal features. The pipeline of STOR can be seen in Figure \ref{fig2}.

\begin{figure}[t]
\centering
\includegraphics[width=8.5cm]{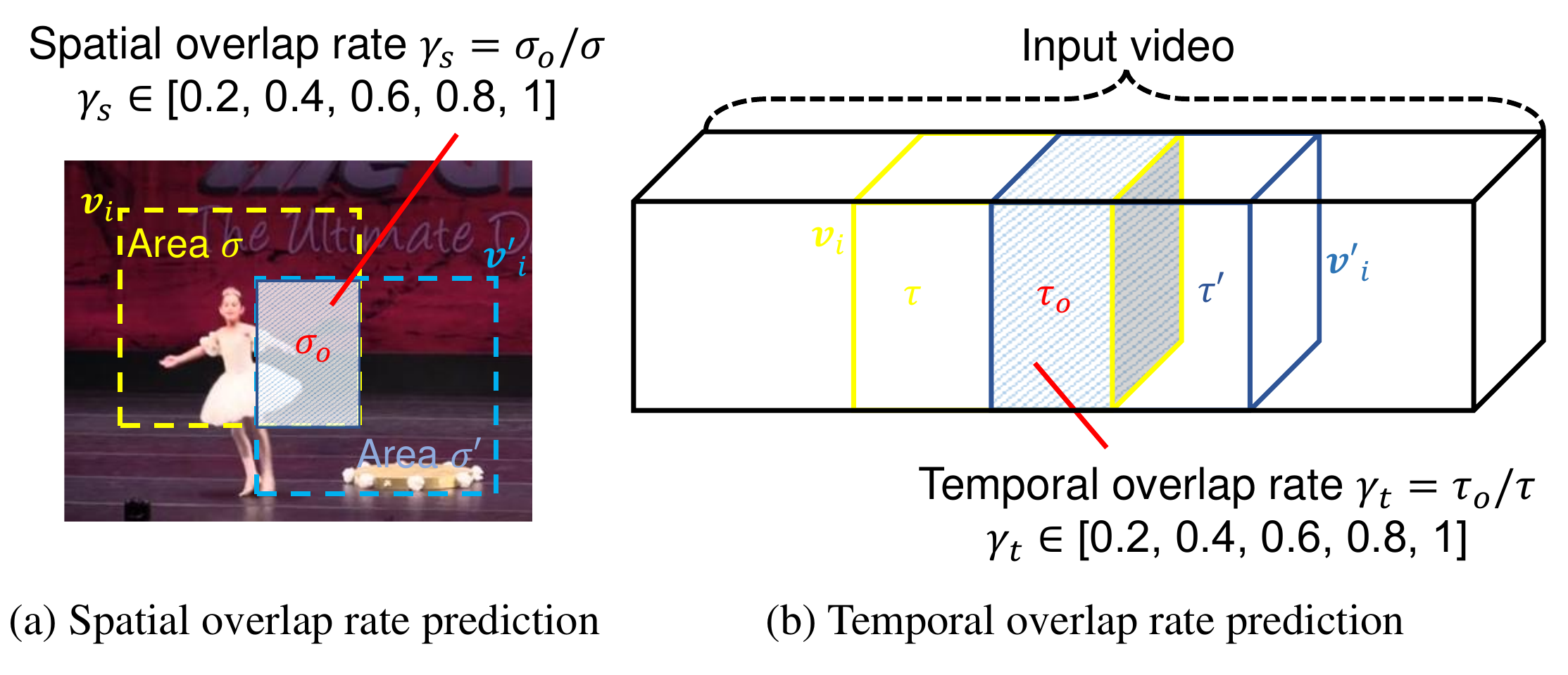}
\caption{Conceptual demonstration of spatio-temporal overlap rate prediction. $\sigma$ denotes the spatial area of bounding box, $\sigma=\sigma'$. $\sigma_o$ denotes spatial overlap area. $\tau$ denotes the temporal overlap duration, $\tau=\tau'$. $\tau_o$ denotes temporal overlap duration.}
\label{fig3}
\end{figure}

The spatio-temporal overlap rate prediction task consists of two parts, i.e., spatial overlap rate prediction and temporal overlap rate prediction. The conceptual demonstration of spatial and temporal overlap rates can be seen in Figure \ref{fig3} (a) and (b). When generating samples $\mathbf{v}_i$ and ${\mathbf{v}'}_i$, spatio-temporal overlap data augmentation will randomly pick a spatial overlap rate $\gamma_s=\frac{\sigma_o}{\sigma}$ and a temporal overlap rate $\gamma_t=\frac{\tau_o}{\tau}$, where $\sigma$ denotes the spatial area of bounding box, $\sigma_o$ is the overlap area, $\tau$ denotes the temporal overlap duration and $\tau_o$ is the temporal overlap duration. Then, it generates two clips with the sampled spatial and temporal overlap rates.

Formally, we denote the STOR as $t_o\sim\mathcal{T}_o$. Given a video $\mathbf{x}_i$ and the neural network $f_\theta$, two random transformations $t_o(\mathbf{x}_i)$ and ${t'}_o(\mathbf{x}_i)$ are applied to obtain the training clip $\mathbf{v}_i, {\mathbf{v}'}_i$, respectively. The randomly sampled spatial overlap rate is $s_o^s$ and the randomly sampled temporal overlap rate is $s_o^t$. Then, we conduct feature extraction and obtain the hidden feature $\mathbf{y}_i=f_\theta(\mathbf{v}_i)$, ${\mathbf{y}'}_i=f_\theta({\mathbf{v}'}_i)$, where the same CNN model $f_\theta$ is used. Furthermore, following the design of multiple tasks for self-supervised learning \cite{chen2020simple}, two individual fully-connected (FC) heads $e_\theta^s$ and $e_\theta^t$ are adopted as the classifiers for spatial overlap rate prediction and temporal overlap rate prediction, respectively. $a_o^s=e_o^s(\mathbf{y}_i)$ and $a_o^t=e_o^t({\mathbf{y}'}_i)$ are used to make a prediction of the sampling label $s_o^s$ and $s_o^t$, respectively. The prediction probabilities are $p_o^s=\frac{\exp{a_o^s}}{\sum_{o}^{N_o^s}\exp{a_o^s}}$, $p_o^t=\frac{\exp{a_o^t}}{\sum_{o}^{N_o^t}\exp{a_o^t}}$, where $N_o^s$ and $N_o^t$ are the number of all the spatial overlap rate and temporal overlap rate candidates, respectively The parameters of the neural network $f_\theta$ are trained with a joint cross entropy (CE) loss $\mathcal{L}_{ocls}$ described as:
\begin{align}
    \mathcal{L}_{ocls}=-l_o^s\sum_{o=1}^{N_o^s}s_o^s\log{p_o^s}-l_o^t\sum_{o=1}^{N_o^t}s_o^t\log{p_o^t},
\end{align}
where $l_o^s$, $l_o^t$ are the weights for spatial and temporal overlap tasks.

\subsection{Spatio-temporal Overlap Data Augmentation}
A spatio-temporal overlap data augmentation is proposed in this work for aforementioned pretext tasks. To implement the proposed spatio-temporal overlap rate prediction task, it is critical to obtain two video clips with overlap rate while keeping the random distribution of the input video clip samples. In the spatio-temporal overlap data augmentation, a 2-step sampling strategy is designed for training. Following previous video self-supervised learning method \cite{wang2021removing, han2020self}, we adopt base data augmentation that includes multi-scale random cropping, random color jittering, random temporal jittering, random rotation jittering (randomly rotate 0¡ã-10¡ã), random playback rates and random rotation (random rotate $0\degree,90\degree,180\degree,270\degree$). Given a video $\mathbf{x}_i$, we generate the first video clip sample $\mathbf{v}_i$ with the base random data augmentation in a video. Then, we randomly pick a spatial overlap rate $s_o^s$ and a temporal overlap rate $s_o^t$, respectively. After that, to keep the same aspect ratio of the first video clip $\mathbf{v}_i$, the second video clip sample ${\mathbf{v}'}_i$ adopts the same spatial size (i.e. width $w$ and height $h$) and the same playback rate of the first video clip sample $\mathbf{v}_i$. Then, the second video clip sample ${\mathbf{v}'}_i$ randomly crops spatial area with size $(w, h)$ according to the picked overlap rate $s_o^s$ and randomly clips temporal duration according to the temporal overlap rate $s_o^t$. After that, base data augmentation except from random cropping and temporal jittering is implemented on ${\mathbf{v}'}_i$.
In this way, the video clip samples $\mathbf{v}_i$ and ${\mathbf{v}'}_i$ have a spatio-temporal overlap rate with each other, meanwhile, follow a random distribution of data augmentation.

\subsection{Full Scheme of CSTP}
\begin{figure*}[t]
\centering
\includegraphics[width=17cm]{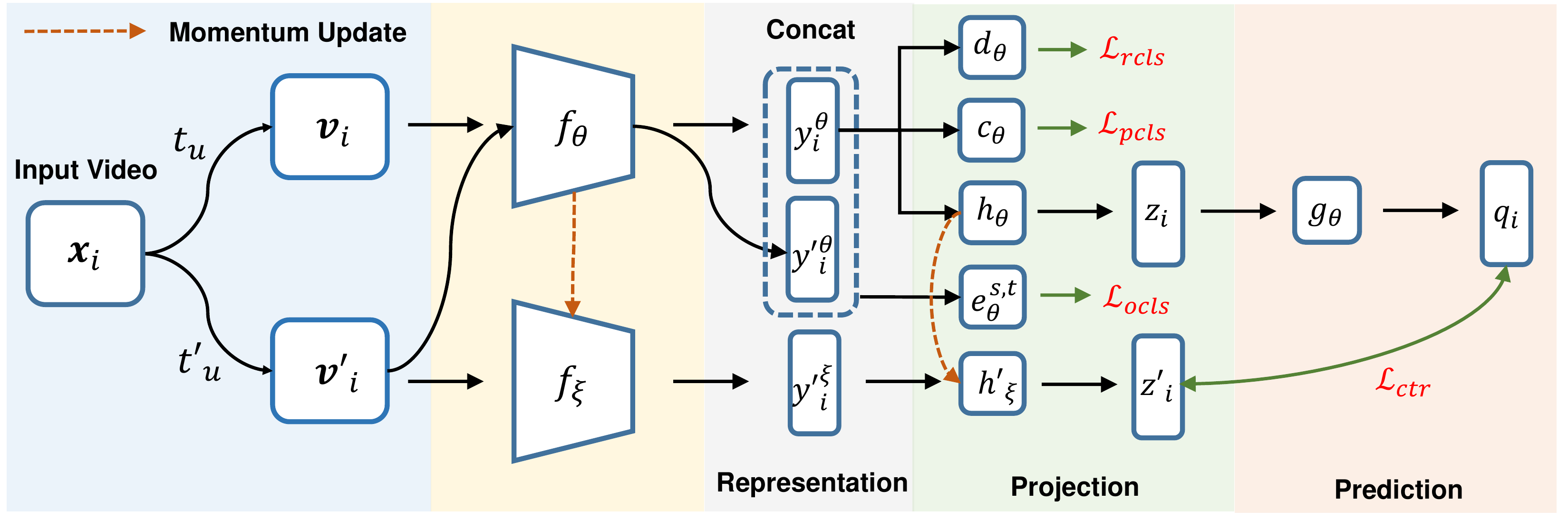}
\caption{An overview of contrastive spatio-temporal pretext (CSTP) approach for video self-supervised learning. The approach consists of contrastive learning and three pretext tasks (playback rate prediction, rotation prediction, spatio-temporal overlap rate (STOR) prediction).}
\label{fig4}
\end{figure*}

Taking BYOL as the contrastive learning method for example, the full framework of our proposed CSTP can be seen in Figure \ref{fig4}. Given an input video $\mathbf{x}_i$, we first randomly generate two fixed-length clips, i.e., $\mathbf{v}_i$ and ${\mathbf{v}'}_i$, from different spatial and temporal locations of $\mathbf{x}_i$ according to the spatio-temporal overlap data augmentation. In this way, the two input video clips $\mathbf{v}_i$ and ${\mathbf{v}'}_i$ have different low-level (pixel, curve, et al.) distribution but are consistent in the semantic level. Then, the two sampled video clips $\mathbf{v}_i$ and ${\mathbf{v}'}_i$ are fed into a 3D CNNs $f_\theta$ and ${f'}_\xi$ to extract the feature representation $y_i^\theta$ and ${y'}_i^\xi$, respectively. Besides, ${\mathbf{v}'}_i$ is fed into $f_\theta$ to extract feature representation ${y'}_i^\theta$. Afterwards, the extracted feature $y_i^\theta$ is fed into playback rate prediction head $c_\theta$ for playback rate prediction and obtain the loss $\mathcal{L}_{pcls}$; The extracted feature $y_i^\theta$ is fed into rotation prediction head $d_\theta$ for rotation prediction and obtain loss the loss $\mathcal{L}_{rcls}$; The extracted features $y_i^\theta$ and ${y'}_i^\theta$ are concatenated and the concatenated features $cat(y_i^\theta,{y'}_i^\theta)$ are fed into $e_\theta^s$ and $e_\theta^t$ for spatial overlap rate prediction and temporal overlap rate prediction, respectively, and obtain the loss $\mathcal{L}_{ocls}$ as Equation (4); The three losses $\mathcal{L}_{pcls}$, $\mathcal{L}_{rcls}$, $\mathcal{L}_{ocls}$ are all CE loss. The extracted features $y_i^\theta$ and ${y'}_i^\theta$ follow the steps of contrastive learning scheme and obtain the loss $\mathcal{L}_{ctr}$. Lastly, contrastive loss and three spatio-temporal pretext loss are jointed for optimization. The final loss is as:
\begin{align}
    \mathcal{L}=\lambda_{ctr}\mathcal{L}_{ctr}+\lambda_{pcls}\mathcal{L}_{pcls}+\lambda_{rcls}\mathcal{L}_{rcls}+\lambda_{ocls}\mathcal{L}_{ocls},
\end{align}
where $\lambda_{ctr}$, $\lambda_{pcls}$, $\lambda_{rcls}$, $\lambda_{ocls}$ are hyperparameters that control the weights of the regularization term.

\section{Experiments}
\subsection{Datasets}
\textbf{Kinetics-400} \cite{carreira2017quo} is one of the large-scale action recognition benchmarks which contains around 300K videos over 400 action categories. \textbf{UCF-101} \cite{soomro2012ucf101} is a widely used benchmark for action recognition. It has three splits, which consists of 13, 320 videos that cover 101 human action classes. \textbf{HMDB-51} \cite{kuehne2011hmdb} is also a small-scale dataset for action recognition, which consists of three splits and 6, 770 videos in 51 actions.

\subsection{Ablations Studies}
\textbf{Understanding of transformation-based pretext tasks.} 3D Rot \cite{jing2018self}, RTT \cite{jenni2020video} has verified that video transformation-based pretext task can benefit video self-supervised learning. It was claimed that the gain on performance comes from the pseudo labels learning, however, we found that the gain in performance of transformation-based pretext task comes from two aspects, i.e., more diverse data (data augmentation along with pretext task) and the pseudo label prediction.

In this study, we adopt R(2+1)D as the backbone and conduct experiments on split 1 of dataset UCF-101. The results are summarized in Table \ref{tb1}. In the table, ``Base'' means basic data augmentation methods which includes multi-scale random cropping, random gaussian blur, random color jittering, random temporal jittering. Exp. 1 denotes the baseline performance, which was trained from scratch (randomly initializing weights) with the aforementioned data augmentation methods. We can observe that when the model was trained from scratch with basic data augmentation methods, the baseline performance is 60.3\% (Exp. 1). After adding playback rate data augmentations, the performance increased to 67.5\% (+7.2\%, Exp. 2). After further combining playback rate prediction pretext task, the performance increased to 77.2 (+9.7\%, Exp. 5). The same trend can be seen in the rotation data augmentation and pretext task (Exp. 3\&7 vs Exp. 1). Besides, we observe that combining the playback rate data augmentation and the rotation data augmentation can benefit the performance (Exp. 2\&3 vs Exp. 4). In terms of our proposed STOR prediction task, it should be noted that unlike playback rates and rotation operation, our proposed overlap operations will not increase the diversity of data. When performing STOR with base data augmentation, the performance increased to 69.6\% (+9.3\%). The gain in performance is comparable to playback rate prediction and rotation prediction, which verifies the effectiveness of our proposed STOR. And we also can observe the further gain with additional data augmentation methods (Exp. 10\&11 vs Exp. 9).

\begin{table}[tb!]
\setlength{\abovecaptionskip}{0mm}
	\renewcommand{\arraystretch}{1.5}
	\caption{Exploration of transformation-based pretext tasks. Results are evaluated on \textbf{split 1} of UCF-101. The backbone is R(2+1)D. ``Base'' means basic data augmentation methods which includes multi-scale random cropping, random gaussian blur, random color jittering, random temporal jittering.}
	\label{tb1}
    \resizebox{\columnwidth}{!}{
    		\begin{tabular}{cccccccc}
    			\hline\hline \\[-5.5mm]
    			\multirow{2}*{Exp.} & \multicolumn{3}{c}{Data Augmentation} & \multicolumn{3}{c}{Pretext Tasks} & Dataset \\
                  & Base & PB Rate & Rot. & PB Rate & Rot. & STOR & UCF-101 (\%) \\
                \hline
                 1 & \textcolor{green}{\Checkmark} & \textcolor{red}{\XSolidBrush} & \textcolor{red}{\XSolidBrush} & \textcolor{red}{\XSolidBrush} & \textcolor{red}{\XSolidBrush} & \textcolor{red}{\XSolidBrush} & 60.3 \\
                 2 & \textcolor{green}{\Checkmark} & \textcolor{green}{\Checkmark} & \textcolor{red}{\XSolidBrush} & \textcolor{red}{\XSolidBrush} & \textcolor{red}{\XSolidBrush} & \textcolor{red}{\XSolidBrush} & 67.5 \\
                 3 & \textcolor{green}{\Checkmark} & \textcolor{red}{\XSolidBrush} & \textcolor{green}{\Checkmark} & \textcolor{red}{\XSolidBrush} & \textcolor{red}{\XSolidBrush} & \textcolor{red}{\XSolidBrush} & 66.9 \\
                 4 & \textcolor{green}{\Checkmark} & \textcolor{green}{\Checkmark} & \textcolor{green}{\Checkmark} & \textcolor{red}{\XSolidBrush} & \textcolor{red}{\XSolidBrush} & \textcolor{red}{\XSolidBrush} & 71.1 \\
                 5 & \textcolor{green}{\Checkmark} & \textcolor{green}{\Checkmark} & \textcolor{red}{\XSolidBrush} & \textcolor{green}{\Checkmark} & \textcolor{red}{\XSolidBrush} & \textcolor{red}{\XSolidBrush} & 77.2 \\
                 6 & \textcolor{green}{\Checkmark} & \textcolor{green}{\Checkmark} & \textcolor{green}{\Checkmark} & \textcolor{green}{\Checkmark} & \textcolor{red}{\XSolidBrush} & \textcolor{red}{\XSolidBrush} & 78.7 \\
                 7 & \textcolor{green}{\Checkmark} & \textcolor{red}{\XSolidBrush} & \textcolor{green}{\Checkmark} & \textcolor{red}{\XSolidBrush} & \textcolor{green}{\Checkmark} & \textcolor{red}{\XSolidBrush} & 77.6 \\
                 8 & \textcolor{green}{\Checkmark} & \textcolor{green}{\Checkmark} & \textcolor{green}{\Checkmark} & \textcolor{red}{\XSolidBrush} & \textcolor{green}{\Checkmark} & \textcolor{red}{\XSolidBrush} & 78.5 \\
                 9 & \textcolor{green}{\Checkmark} & \textcolor{red}{\XSolidBrush} & \textcolor{red}{\XSolidBrush} & \textcolor{red}{\XSolidBrush} & \textcolor{red}{\XSolidBrush} & \textcolor{green}{\Checkmark} & 69.6 \\
                 10 & \textcolor{green}{\Checkmark} & \textcolor{green}{\Checkmark} & \textcolor{red}{\XSolidBrush} & \textcolor{red}{\XSolidBrush} & \textcolor{red}{\XSolidBrush} & \textcolor{green}{\Checkmark} & 74.1 \\
                 11 & \textcolor{green}{\Checkmark} & \textcolor{green}{\Checkmark} & \textcolor{green}{\Checkmark} & \textcolor{red}{\XSolidBrush} & \textcolor{red}{\XSolidBrush} & \textcolor{green}{\Checkmark} & \textbf{76.2} \\
    			\hline\hline
    		\end{tabular}
     }
\end{table}

\textbf{Exploration of CSTP Learning Methods.} To verify the effectiveness of our proposed CSTP and figure out the impact of each contribution, experiments on each part of CSTP were conducted. We take BYOL as the contrastive learning framework for exploration. R(2+1)D was adopted as the backbone in the experiments. The performance of each part in CSTP is shown in Table \ref{tb2}. Firstly, we can observe that each part of CSTP can boost the performance of the baseline. The baseline performance is 71.1\% (scratch). The performance was further increased to 76.2\% (+5.1\%, Exp. 2), 78.7\% (+7.6\%, Exp. 3), 78.5\% (+7.4\%, Exp. 4), 76.6\% (+5.5\%, Exp. 5) when the training was conducted from the pre-trained model of contrastive learning, playback rate prediction, rotation prediction and overlap rate prediction, respectively. Secondly, it can be observed that combining each pretext task to contrastive learning can boost the performance of individually using contrastive learning or the pretext task. Thirdly, STOR can boost the performance of each component in CSTP. When STOR is combined with playback rate prediction and rotation prediction, the performance is improved by +3.3\% and +3.1\%, respectively. When combining STOR with contrastive learning, the performance is improved by 1.1\%. The reason why the improvement on contrastive learning is less than pretext task may be that STOR and contrastive learning both are learning the similarity of samples, which is independent to playback rate prediction and rotation prediction. Finally, through comparisons (Exp. 6 vs Exp. 3\&4; Exp. 7 vs Exp. 3\&5, etc.), it can be observed that in CSTP, the combination of multiple pretext tasks and contrastive learning can improve the overall performance, which verifies the importance of the learned intermediate state in representation.

\begin{table}[tb!]
\setlength{\abovecaptionskip}{0mm}
	\renewcommand{\arraystretch}{1.5}
	\caption{Exploration of each part of contrastive spatio-temporal pretext learning methods. Results are evaluated on \textbf{split 1} of UCF-101 and HMDB-51.}
	\label{tb2}
	\centering
    \resizebox{7.5cm}{!}{
    		\begin{tabular}{ccccccc}
    			\hline\hline \\[-5.5mm]
    			\multirow{2}{*}{Exp.} & \multirow{2}*{Contr.} & \multicolumn{3}{c}{Pretext Tasks} & \multicolumn{2}{c}{Dataset (\%)} \\
                  & & PB Rate & Rot. & STOR & UCF-101 & HMDB-51 \\
                \hline
                 1 & \textcolor{red}{\XSolidBrush} & \textcolor{red}{\XSolidBrush} & \textcolor{red}{\XSolidBrush} & \textcolor{red}{\XSolidBrush} & 71.1 & 38.3 \\
                 2 & \textcolor{green}{\Checkmark} & \textcolor{red}{\XSolidBrush} & \textcolor{red}{\XSolidBrush} & \textcolor{red}{\XSolidBrush} & 76.6 & 45.5 \\
                 3 & \textcolor{red}{\XSolidBrush} & \textcolor{green}{\Checkmark} & \textcolor{red}{\XSolidBrush} & \textcolor{red}{\XSolidBrush} & 78.7 & 48.2 \\
                 4 & \textcolor{red}{\XSolidBrush} & \textcolor{red}{\XSolidBrush} & \textcolor{green}{\Checkmark} & \textcolor{red}{\XSolidBrush} & 78.5 & 47.1 \\
                 5 & \textcolor{red}{\XSolidBrush} & \textcolor{red}{\XSolidBrush} & \textcolor{red}{\XSolidBrush} & \textcolor{green}{\Checkmark} & 76.2 & 45.4 \\
                \hline
                 6 & \textcolor{red}{\XSolidBrush} & \textcolor{green}{\Checkmark} & \textcolor{green}{\Checkmark} & \textcolor{red}{\XSolidBrush} & 84.1 & 53.5 \\
                 7 & \textcolor{red}{\XSolidBrush} & \textcolor{green}{\Checkmark} & \textcolor{red}{\XSolidBrush} & \textcolor{green}{\Checkmark} & 82.0 & 51.7 \\
                 8 & \textcolor{red}{\XSolidBrush} & \textcolor{red}{\XSolidBrush} & \textcolor{green}{\Checkmark} & \textcolor{green}{\Checkmark} & 81.6 & 51.4 \\
                 9 & \textcolor{green}{\Checkmark} & \textcolor{green}{\Checkmark} & \textcolor{red}{\XSolidBrush} & \textcolor{red}{\XSolidBrush} & 81.8 & 51.4 \\
                 10 & \textcolor{green}{\Checkmark} & \textcolor{red}{\XSolidBrush} & \textcolor{green}{\Checkmark} & \textcolor{red}{\XSolidBrush} & 81.5 & 51.1 \\
                 11 & \textcolor{green}{\Checkmark} & \textcolor{red}{\XSolidBrush} & \textcolor{red}{\XSolidBrush} & \textcolor{green}{\Checkmark} & 77.7 & 49.5 \\
                \hline
                 12 & \textcolor{red}{\XSolidBrush} & \textcolor{green}{\Checkmark} & \textcolor{green}{\Checkmark} & \textcolor{green}{\Checkmark} & 85.3 & 54.9 \\
                 13 & \textcolor{green}{\Checkmark} & \textcolor{green}{\Checkmark} & \textcolor{green}{\Checkmark} & \textcolor{red}{\XSolidBrush} & 84.4 & 54.1 \\
                 14 & \textcolor{green}{\Checkmark} & \textcolor{green}{\Checkmark} & \textcolor{red}{\XSolidBrush} & \textcolor{green}{\Checkmark} & 83.1 & 53.8 \\
                 15 & \textcolor{green}{\Checkmark} & \textcolor{red}{\XSolidBrush} & \textcolor{green}{\Checkmark} & \textcolor{green}{\Checkmark} & 82.6 & 52.7 \\
                \hline
                 16 & \textcolor{green}{\Checkmark} & \textcolor{green}{\Checkmark} & \textcolor{green}{\Checkmark} & \textcolor{green}{\Checkmark} & \textbf{85.6} & \textbf{55.1} \\
    			\hline\hline
    		\end{tabular}
     }
\end{table}

\textbf{Exploration of Contrastive Learning Methods.}
To explore the mutual influence of multiple contrastive learning schemes and different pretext tasks, we conducted experiments on four popular contrastive learning schemes which are SimCLR, MoCo, BYOL and SimSiam. In Table \ref{tb3}, we compared four contrastive learning schemes in the context of CSTP. Except from the contrastive learning scheme, the settings of the experiments in Table \ref{tb3} are the same as those in Table \ref{tb2}. Firstly, we observe that four contrastive learning schemes can work with the pretext tasks in CSTP by comparing the results of each contrastive learning schemes in Table \ref{tb3} with the results of only pretext tasks in Table \ref{tb2}. For example, when only using SimCLR in CSTP, the performance is 75.5\%, while the performance increased to 80.7\% (+5.2\%), 82.9\% (+7.4\%) and 84.4\% (+8.9\%) as gradually combining playback rate prediction, rotation prediction and STOR prediction. Secondly, in the four contrastive learning schemes, we observe that BYOL performed the best in our proposed CSTP framework. Thirdly, we observe the four contrastive learning schemes share a similar trend when combining different pretext tasks. Besides, we observe that the four contrastive learning schemes share a similar trend when combining with pretext tasks. Finally, we can observe that momentum update is useful for CSTP (comparing performance of BYOL and SiaSiam, BYOL outperforms SiaSiam); negative samples are not critical for CSTP when using momentum update (comparing performance of BYOL and MoCo, BYOL outperforms MoCo).
\begin{table}[tb!]
\setlength{\abovecaptionskip}{0mm}
	\renewcommand{\arraystretch}{1.5}
	\caption{Exploration of contrastive learning methods of CSTP. Results are evaluated on \textbf{split} 1 of UCF-101 and HMDB-51.}
	\label{tb3}
	\centering
    \resizebox{6.5cm}{!}{
    		\begin{tabular}{cccccc}
    			\hline\hline \\[-5.5mm]
    			\multirow{2}{*}{Contr.} & \multicolumn{3}{c}{Pretext Tasks} & \multicolumn{2}{c}{Dataset (\%)} \\
                  & PB Rate & Rot. & STOR & UCF-101 & HMDB-51 \\
                \hline
                \multirow{4}{*}{SimCLR} & \textcolor{red}{\XSolidBrush} & \textcolor{red}{\XSolidBrush} & \textcolor{red}{\XSolidBrush} & 75.5 & 44.3 \\
                                        & \textcolor{green}{\Checkmark} & \textcolor{red}{\XSolidBrush} & \textcolor{red}{\XSolidBrush} & 80.7 & 50.1 \\
                                        & \textcolor{green}{\Checkmark} & \textcolor{green}{\Checkmark} & \textcolor{red}{\XSolidBrush} & 82.9 & 52.8 \\
                                        & \textcolor{green}{\Checkmark} & \textcolor{green}{\Checkmark} & \textcolor{green}{\Checkmark} & 84.4 & 54.0 \\
                \hline
                \multirow{4}{*}{MoCo} & \textcolor{red}{\XSolidBrush} & \textcolor{red}{\XSolidBrush} & \textcolor{red}{\XSolidBrush} & 75.8 & 44.8 \\
                                      & \textcolor{green}{\Checkmark} & \textcolor{red}{\XSolidBrush} & \textcolor{red}{\XSolidBrush} & 81.3 & 52.8 \\
                                      & \textcolor{green}{\Checkmark} & \textcolor{green}{\Checkmark} & \textcolor{red}{\XSolidBrush} & 83.5 & 53.4 \\
                                      & \textcolor{green}{\Checkmark} & \textcolor{green}{\Checkmark} & \textcolor{green}{\Checkmark} & 84.9 & 54.3 \\
                \hline
                \multirow{4}{*}{BYOL} & \textcolor{red}{\XSolidBrush} & \textcolor{red}{\XSolidBrush} & \textcolor{red}{\XSolidBrush} & 76.6 & 45.5 \\
                                      & \textcolor{green}{\Checkmark} & \textcolor{red}{\XSolidBrush} & \textcolor{red}{\XSolidBrush} & 81.8 & 51.4 \\
                                      & \textcolor{green}{\Checkmark} & \textcolor{green}{\Checkmark} & \textcolor{red}{\XSolidBrush} & 84.4 & 54.1 \\
                                      & \textcolor{green}{\Checkmark} & \textcolor{green}{\Checkmark} & \textcolor{green}{\Checkmark} & \textbf{85.6} & \textbf{55.1} \\
                \hline
                \multirow{4}{*}{SimSiam} & \textcolor{red}{\XSolidBrush} & \textcolor{red}{\XSolidBrush} & \textcolor{red}{\XSolidBrush} & 75.2 & 44.2 \\
                                         & \textcolor{green}{\Checkmark} & \textcolor{red}{\XSolidBrush} & \textcolor{red}{\XSolidBrush} & 80.6 & 50.3 \\
                                         & \textcolor{green}{\Checkmark} & \textcolor{green}{\Checkmark} & \textcolor{red}{\XSolidBrush} & 82.8 & 52.6 \\
                                         & \textcolor{green}{\Checkmark} & \textcolor{green}{\Checkmark} & \textcolor{green}{\Checkmark} & 83.9 & 53.6 \\
    			\hline\hline
    		\end{tabular}
     }
\end{table}

\textbf{Exploration of STOR.}
From Table \ref{tb2}\&\ref{tb3}, we observe that combining STOR benefits the overall results in CSTP. STOR consists of two pretext tasks, e.g., spatial overlap rate prediction and temporal overlap prediction. To demonstrate the effectiveness of these two pretext tasks, we separate STOR apart and conduct experiments on each part. To remove the influence of the data augmentation methods of other pretext tasks, "Base" data augmentation is only used in the experiments. The results are summarized in Table \ref{tb4}. When training from scratch, the result in UCF-101 is 60.3\%. While after employing spatial overlap prediction pretext task, the performance increased to 67.8\% (+7.5\%). Similar trend can be observed in temporal overlap prediction, which increased the baseline to 66.5\% (+6.2\%). Besides, combining the spatial overlap prediction and temporal overlap prediction can further increase the performance to 69.6\% (+9.3\%).

\textbf{Exploration of the choice of different spatio-temporal overlap rates in STOR.}
In this experiment, we explore the influence of different candidates to STOR pre-training strategy. R(2+1)D was adopted as the backbone. We only use STOR as the pre-training strategy and fine-tune the whole model as demonstrated in Experimental Settings in Appendix. We conducted 6 sets of candidates to demonstrate the influence of the choice of candidates, which are 2 candidates - $[0.5, 1]$, 3 candidates - $[0.33, 0.66, 0.99]$, 4 candidates - $[0.25, 0.5, 0.75, 1.0]$, 5 candidates - $[0.2, 0.4, 0.6, 0.8, 1.0]$, 6 candidates - $[0.166, 0.332, 0.498, 0.664, 0.83, 1.0]$, 7 candidates - $[0.143, 0.286, 0.429, 0.572, 0.715, 0.858, 1]$. The comparative results of different candidates of STOR prediction task in UCF-101 and HMDB-51 datasets are shown in Table \ref{tb_candi}.

We can observe that the two candidates in STOR performed the worst results. And with the number of candidates increased, the results tend to improve till the number of candidates reaches five. When the number of candidates is bigger than five, the performance tends to be saturated. The reason may be that few candidates make perceiving STOR prediction easy, so that the model cannot learn spatial-temporal representation well. When increasing the number of candidates, the task of STOR prediction is getting harder, which requires the model stronger representation ability. And when the number of candidates was over five, the STOR task may be too hard to obtain better representation ability.

\begin{table}[tb!]
\setlength{\abovecaptionskip}{0mm}
	\renewcommand{\arraystretch}{1.5}
	\caption{Exploration of the choice of different spatio-temporal overlap rates in STOR. Results are evaluated on \textbf{split 1} of UCF-101 and HMDB-51.}
	\label{tb_candi}
	\centering
    \resizebox{6.5cm}{!}{
    		\begin{tabular}{ccc}
    			\hline\hline \\[-5.5mm]
    			Number of candidates & UCF-101 (\%) & HMDB-51 (\%) \\
                \hline
                2 & 72.5 & 42.5 \\
                3 & 72.4 & 43.3 \\
                4 & 73.9 & 44.8 \\
                5 & \textbf{76.2} & \textbf{45.4} \\
                6 & 75.4 & 45.0 \\
                7 & 75.5 & 44.9 \\
    			\hline\hline
    		\end{tabular}
     }
\end{table}

\begin{table}[tb!]
\setlength{\abovecaptionskip}{0mm}
	\renewcommand{\arraystretch}{1.5}
	\caption{Exploration of STOR. Results are evaluated on \textbf{split 1} of UCF-101. Results without data augmentation methods of other pretext tasks.}
	\label{tb4}
    \resizebox{\columnwidth}{!}{
    		\begin{tabular}{ccc}
    			\hline\hline \\[-5.5mm]
    			Spatial Overlap Prediction & Temporal Overlap Prediction & UCF-101 (\%) \\
                \hline
                 \textcolor{red}{\XSolidBrush} & \textcolor{red}{\XSolidBrush} & 60.3 (baseline) \\
                 \textcolor{green}{\Checkmark} & \textcolor{red}{\XSolidBrush} & 67.8 \\
                 \textcolor{red}{\XSolidBrush} & \textcolor{green}{\Checkmark} & 66.5 \\
                 \textcolor{green}{\Checkmark} & \textcolor{green}{\Checkmark} & \textbf{69.6} \\
    			\hline\hline
    		\end{tabular}
     }
\end{table}

\subsection{Evaluation on Action Recognition Task}
\textbf{End-to-end fine-tuning stage.} Following previous work \cite{chen2021rspnet, jenni2020video}, we compare our proposed method with recent state-of-the-art self-supervised video representation learning approaches. In Table \ref{tb5}, for fair comparison, we report the Top-1 accuracy of UCF-101 and HMDB-51 datasets with commonly used network backbones (C3D, R(2+1)D, S3D) and pre-training datasets, where the results are averaged over three splits of the two datasets. When trained from pre-trained model on UCF-101, it can be observed that our proposed method can achieve state-of-the-art performance among all the approaches with all the backbones. Specifically, our CSTP achieves 85.7\%/55.3\% with R(2+1)D backbone, which outperforms state-of-the-art performance by 4.1\%/6.1\% in UCF-101/HMDB-51 datasets. It even outperforms some approaches trained on Kinetic-400. Similar trend can be observed on C3D and S3D. This demonstrates that our proposed method is generalized to multiple network architectures, which has a robust spatio-temporal modeling ability. When trained from pre-trained model on Kinetics-400, we observe that our CSTP could achieve the state-of-the-art performance in on all the backbones with same input modality. Notably, CSTP unsupervised pre-training leads the accuracy by 0.6\%/1.6\% against fully-supervised ImageNet pre-training with only pre-trained on the small dataset UCF-101.
\begin{table}[tb!]
\setlength{\abovecaptionskip}{0mm}
	\renewcommand{\arraystretch}{1.5}
	\caption{Comparison With State-of-the-arts. Average results of \textbf{three splits} in UCF-101 and HMDB-51 datasets.}
	\label{tb5}
	\centering
    \resizebox{8.5cm}{!}{
        \begin{threeparttable}
    		\begin{tabular}{cccccc}
    			\hline\hline \\[-5.5mm]
                \multirow{2}{*}{Method} & \multirow{2}{*}{Backbone} & \multirow{2}{*}{Dataset} & \multirow{2}{*}{Freeze} & \multicolumn{2}{c}{Datasets (\%)} \\
                 & & & & UCF & HMDB \\
                \hline
                CCL \cite{kong2020cycle} & R3D-18 & K400 & \textcolor{green}{\Checkmark} & 54.0 & 29.5 \\
                MemDPC \cite{han2020memory} & R3D-34 & K400 & \textcolor{green}{\Checkmark} & 54.1 & 30.5 \\
                TaCo \cite{bai2020can} & R3D-18 & K400 & \textcolor{green}{\Checkmark} & 59.6 & 26.7 \\
                MFO \cite{qian2021enhancing} & R3D-18 & K400 & \textcolor{green}{\Checkmark} & 63.2 & 33.4 \\
                Ours & R3D-18 & K400 & \textcolor{green}{\Checkmark} & \textbf{70.5} & \textbf{34.4} \\
                \hline
                \multicolumn{1}{c}{Random Initialization (scratch)} & R(2+1)D & - & \textcolor{red}{\XSolidBrush} & 60.3 & 27.6 \\
                \multicolumn{1}{c}{Supervised ImageNet Pre-training} & R(2+1)D & ImageNet & \textcolor{red}{\XSolidBrush} & 85.1 & 53.7 \\
                \hline
                VCP \cite{luo2020video} & R(2+1)D & UCF-101 & \textcolor{red}{\XSolidBrush} & 66.3 & 32.2 \\
                PRP \cite{yao2020video} & R(2+1)D & UCF-101 & \textcolor{red}{\XSolidBrush} & 72.1 & 35.0 \\
                VCOP \cite{xu2019self} & R(2+1)D & UCF-101 & \textcolor{red}{\XSolidBrush} & 72.4 & 30.9 \\
                PacePred \cite{wang2020self} & R(2+1)D & UCF-101 & \textcolor{red}{\XSolidBrush} & 75.9 & 35.9 \\
                RSPNet \cite{chen2021rspnet} & R(2+1)D & K400 & \textcolor{red}{\XSolidBrush} & 81.1 & 44.6 \\
                RTT \cite{jenni2020video} & R(2+1)D & UCF-101 & \textcolor{red}{\XSolidBrush} & 81.6 & 46.4 \\
                VideoMoCo \cite{pan2021videomoco} & R(2+1)D & K400 & \textcolor{red}{\XSolidBrush} & 78.7 & 49.2 \\
                Ours & R(2+1)D & UCF-101 & \textcolor{red}{\XSolidBrush} & 85.7 & 55.3 \\
                Ours & R(2+1)D & K400 & \textcolor{red}{\XSolidBrush} & \textbf{87.6} & \textbf{56.4} \\
                \hline
                RTT \cite{jenni2020video} & C3D & UCF-101 & \textcolor{red}{\XSolidBrush} & 68.3 & 38.4 \\
                PRP \cite{yao2020video} & C3D & UCF-101 & \textcolor{red}{\XSolidBrush} & 69.1 & 34.5 \\
                VCP \cite{luo2020video} & C3D & UCF-101 & \textcolor{red}{\XSolidBrush} & 68.5 & 32.5 \\
                VCOP \cite{xu2019self} & C3D & UCF-101 & \textcolor{red}{\XSolidBrush} & 65.6 & 28.4 \\
                MoCo+BE \cite{wang2021removing} & C3D & UCF-101 & \textcolor{red}{\XSolidBrush} & 72.4 & 42.3 \\
                RSPNet \cite{chen2021rspnet} & C3D & K400 & \textcolor{red}{\XSolidBrush} & 76.7 & 44.6 \\
                Ours & C3D & UCF-101 & \textcolor{red}{\XSolidBrush} & 80.8 & 48.9 \\
                Ours & C3D & K400 & \textcolor{red}{\XSolidBrush} & \textbf{82.6} & \textbf{50.2} \\
                \hline
                MFO \cite{qian2021enhancing} & S3D & UCF-101 & \textcolor{red}{\XSolidBrush} & 74.3 & 37.2 \\
                CoCLR* \cite{han2020self} & S3D & UCF-101 & \textcolor{red}{\XSolidBrush} & 81.4 & 52.1 \\
                CoCLR* \cite{han2020self} & S3D & K400 & \textcolor{red}{\XSolidBrush} & \textbf{87.9} & 54.6 \\
                Ours & S3D & UCF-101 & \textcolor{red}{\XSolidBrush} & 83.6 & 53.3 \\
                Ours & S3D & K400 & \textcolor{red}{\XSolidBrush} & 87.3 & \textbf{55.7} \\
    			\hline\hline
    		\end{tabular}
            \begin{tablenotes}
                \item *CoCLR used optical flow as guidance on training.
            \end{tablenotes}
        \end{threeparttable}
     }
\end{table}

\textbf{Linear Evaluation.}
We also follow the settings of previous work \cite{jenni2020video} for the linear evaluation in which we freeze the convolutional layers and only train the final FC layers for classification. The ¡°Freeze¡± column in Table \ref{tb5} denotes the linear evaluation setting. We can observe that our method obtains state-of-the-arts performance on UCF-101 and HMDB-51. Specifically, our approach also outperforms TaCo \cite{bai2020can} which combined four existing pretexts tasks with MoCo, demonstrating the representation ability of our proposed method.

\subsection{Evaluation on Video Retrieval Task}
Besides the video action recognition task, we also report the Top-$k$ ($k=1, 5, 10, 20, 50$) video retrieval performance with two different backbones, e.g., R(2+1)D and C3D. The quantitative results are shown in Table \ref{tb6}. We observe that our proposed approach outperforms state-of-the-art approaches by a large margin under different $k$ with the two network backbones. Specifically, our CSTP outperforms the state-of-the-art model by 16.6\%/6.9\% in UCF-101/HMDB-51 dataset using R(2+1)D as the backbone, and outperforms the state-of-the-art model by 14.7\%/6.7\% using C3D as the backbone. These imply that our proposed CSTP could help the model to learn more discriminating spatio-temporal features.

\begin{table}[tb!]
\setlength{\abovecaptionskip}{0mm}
	\renewcommand{\arraystretch}{1.5}
	\caption{Recall-at-Topk. Comparison with state-of-the-art methods in video retrieval task on UCF-101 and HMDB-51.}
	\label{tb6}
	\centering
    \resizebox{8.5cm}{!}{
        \begin{threeparttable}
    		\begin{tabular}{cccccccc}
    			\hline\hline
                \multirow{2}{*}{Method} & \multirow{2}{*}{Backbone} & \multicolumn{3}{c}{UCF-101 (\%)} & \multicolumn{3}{c}{HMDB-51 (\%)} \\
                 & & R@1 & R@5 & R@10 & R@1 & R@5 & R@10 \\
                 RSPNet [2021] & R3D-18 & 41.1 & 59.4 & 68.4 & - & - & - \\
                 MFO [2021] & R3D-18 & 39.6 & 57.6 & 69.2 & 18.8 & 39.2 & 51.0 \\
                 \hline
                 VCOP [2019] & R(2+1)D & 10.7 & 25.9 & 35.4 & 5.7 & 19.5 & 30.7  \\
                 VCP [2020] & R(2+1)D & 19.9 & 33.7 & 42.0 & 6.7 & 21.3 & 32.7 \\
                 PRP [2020] & R(2+1)D & 20.3 & 34.0 & 41.9 & 8.2 & 25.3 & 36.2 \\
                 PacePred [2020] & R(2+1)D & 25.6 & 42.7 & 51.3 & 12.9 & 31.6 & 43.2 \\
                 IIC [2020] & R(2+1)D & 34.7 & 51.7 & 60.9 & 12.7 & 33.3 & 45.8 \\
                 Ours & R(2+1)D & \textbf{51.3} & \textbf{71.1} & \textbf{80.3} & \textbf{21.6} & \textbf{48.4} & \textbf{62.4} \\
                 \hline
                 VCOP [2019] & C3D & 12.5 & 29.0 & 39.0 & 7.4 & 22.6 & 34.4 \\
                 VCP [2020] & C3D & 17.3 & 31.5 & 42.0 & 7.8 & 23.8 & 35.3 \\
                 PRP [2020] & C3D & 23.2 & 38.1 & 46.0 & 10.5 & 27.2 & 40.4 \\
                 PacePred [2020] & C3D & 31.9 & 49.7 & 59.2 & 12.5 & 32.2 & 45.4 \\
                 IIC [2020] & C3D & 31.9 & 48.2 & 57.3 & 11.5 & 31.3 & 43.9 \\
                 RSPNet [2021] & C3D & 36.0 & 56.7 & 66.5 & - & - & - \\
                 MoCo+BE [2021] & C3D & - & - & - & 10.2 & 27.6 & 40.5 \\
                 Ours & C3D & \textbf{50.7} & \textbf{69.4} & \textbf{77.9} & \textbf{19.3} & \textbf{46.6} & \textbf{61.0} \\
    			\hline\hline
    		\end{tabular}
            \begin{tablenotes}
                \item Retrieval results of Top-50 are shown in Appendix.
            \end{tablenotes}
         \end{threeparttable}
     }
\end{table}

\subsection{Qualitative analysis}
\begin{figure}[t]
\centering
\includegraphics[width=8.5cm]{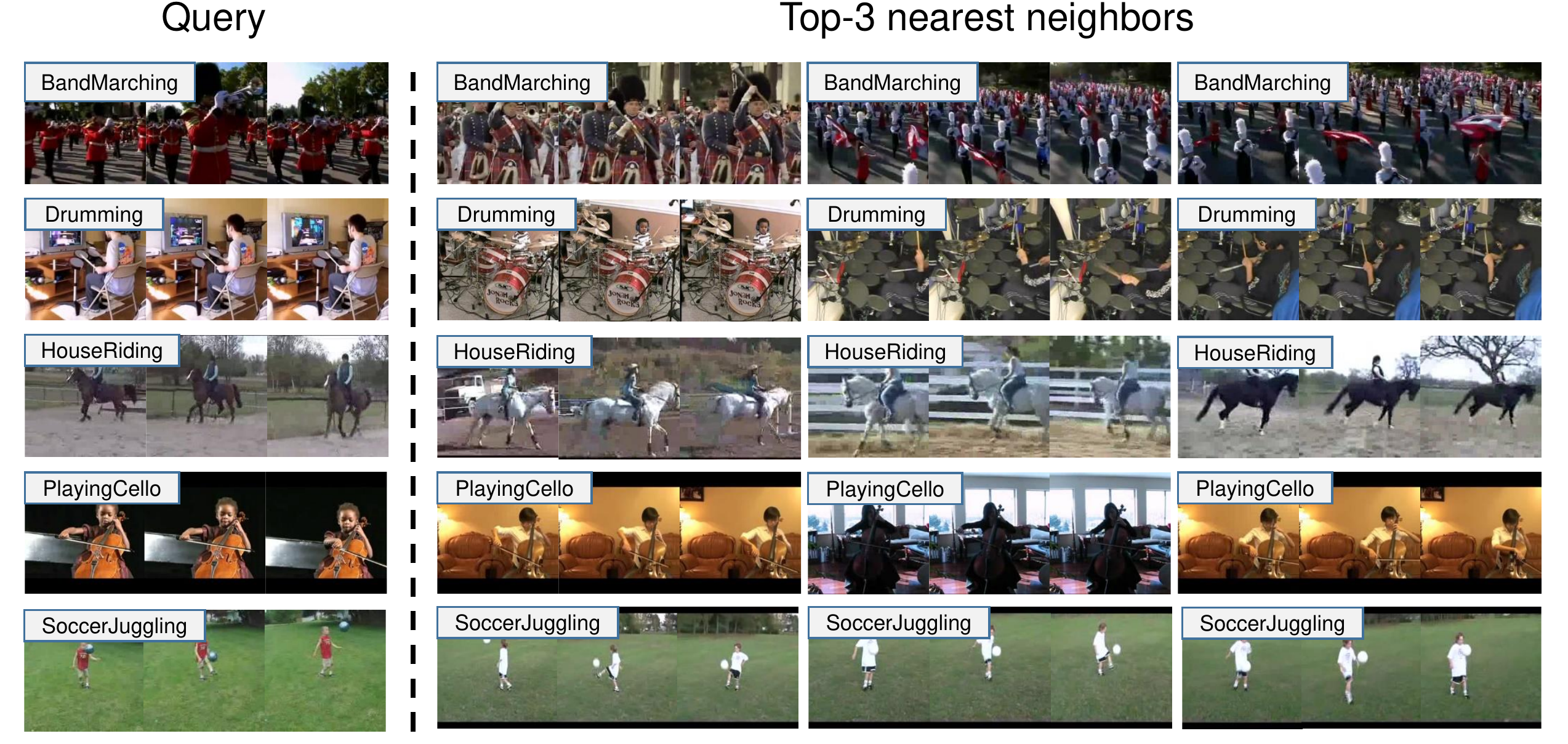}
\caption{Qualitative examples of video retrieval. Retrieval of Top-3 similar samples.}
\label{fig_retri}
\end{figure}

\begin{figure}[t]
\centering
\includegraphics[width=8.5cm]{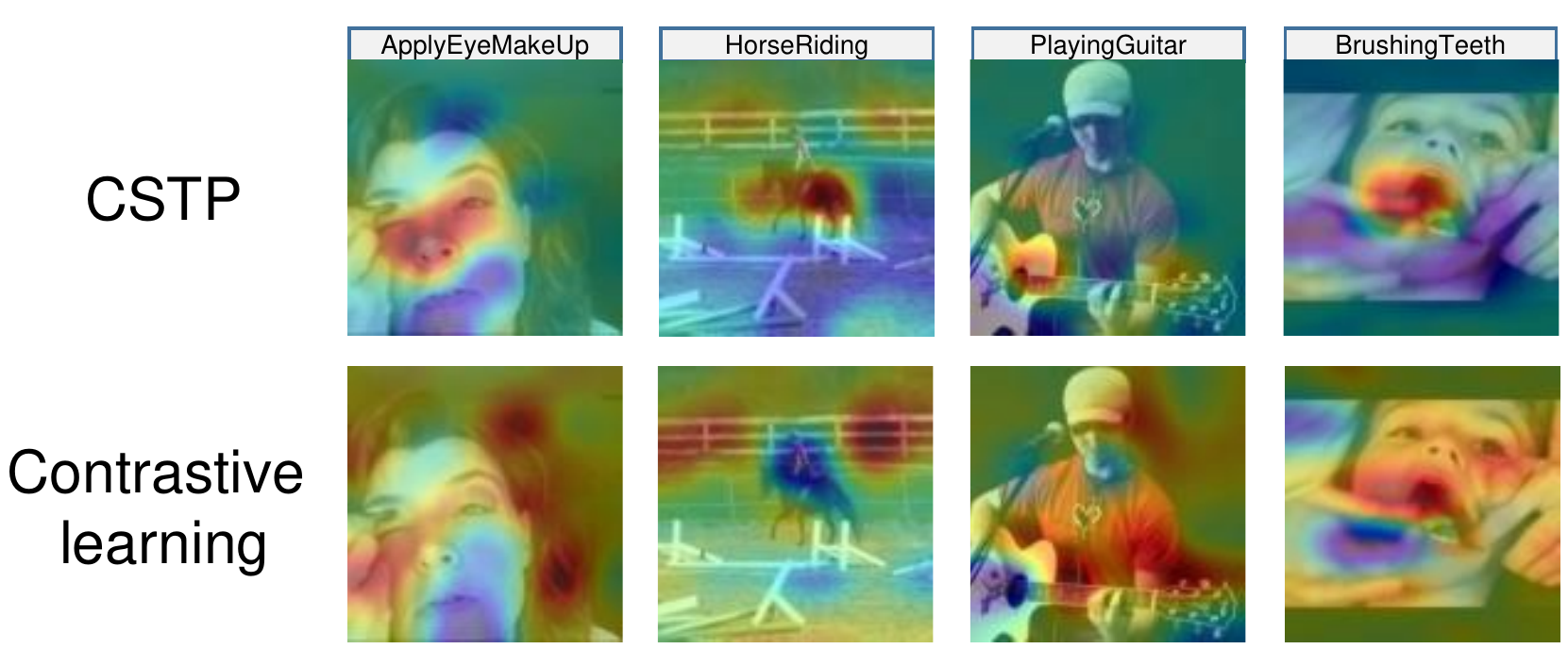}
\caption{Visualization of RoI learn for contrastive learning and CSTP. Our model focuses more on the regions containing motion and appearance information for two pretext tasks, respectively.}
\label{fig_cam}
\end{figure}

We further provide retrieval results as a qualitative study for the proposed CSTP as shown in Figure \ref{fig_retri}. We can see that for the listed five query clips, videos with similar motion characteristics and appearance are successfully retrieved. This implies that CSTP can learn both meaningful motion and appearance representations for videos.
To better understand the learned clues of CSTP, we use the class activation map technique (CAM) \cite{zhou2016learning} to visualize the Region of Interest (RoI) following the previous works \cite{qian2021enhancing, kuang2021video}. We visualize the activations of the output of the last convolutional layers from R(2+1)D models pretrained by contrastive learning (BYOL) and the proposed CSTP (BYOL based). As seen in Figure \ref{fig_cam}, CSTP can focus more on discriminative motion regions, while contrastive learning cannot perceive important motion cues well. For example, our approach precisely focuses on the moving hands in the playing guitar scene, while contrastive learning approach regards the body and some background as the RoI.

\vspace{-5mm}

\section{Conclusion}
Contrastive learning focuses on discriminating the similarity of features while ignoring the intermediate state of learned representations. In this paper, considering the degree of similarity as the intermediate state, we propose a new pretext task - spatio-temporal overlap rate prediction (STOR) in a way of inter-feature reference, which enhances spatio-temporal representation learning by discriminating the STOR of two relative samples. Besides, we propose a joint optimization framework contrastive spatio-temporal pretext (CSTP) to further enhance spatio-temporal feature learning. Furthermore, we study the mutual influence of each component in CSTP and provide design guidance. Extensive experiments show that the STOR prediction task can benefit self-supervised learning. Besides, CSTP framework is flexible and achieves state-of-the-art performance on action recognition and video retrieval downstream tasks with different backbones. It is worth noting that the way of intermediate state perception and inter-feature reference in STOR provides new perspectives for self-supervised learning community.

\bibliography{aaai22}

\clearpage

\section{Appendix}
\subsection{Experimental Settings}
In common practice for video self-supervised model evaluation, researchers normally adopt two evaluation protocols, i.e., supervised fine-tuning and video retrieval. Following \cite{luo2020video, xu2019self}, we also adopt these two protocols for evaluation of our proposed approach.

\textbf{Pre-training details.} Recent video self-supervised learning methods used multiple 3D CNN backbones for evaluation. Following \cite{jenni2020video, wang2020self}, we conducted experiments on famous backbones in action recognition, C3D \cite{tran2015learning}, R(2+1)D \cite{tran2018closer}, S3D \cite{xie2018rethinking} in this work for evaluation of our proposed method. In ablation studies in this paper, we mainly adopt R(2+1)D as the backbone for exploration. In the pre-training stage, following prior works \cite{chen2020simple, he2020momentum}, we use a MLP head after the feature representation for projection. The head takes the global pooling feature as the input and embeds the feature into 128 dimensions with two fully-connected layers ($n\times4096$ and $4096\times128$, $n$ is the dimension of output feature representation). Note that, the MLP heads were only used in the pre-training stage and not involved in downstream tasks. The output feature of the MLP is normalized for feature comparison. The batch size in our implementations was 60 for UCF-101 dataset and 128 for Kinetics-400. Following \cite{grill2020bootstrap}, the momentum coefficient $m$ for BYOL adopts 0.998. Following \cite{chen2020simple, he2020momentum}, the temperature $\alpha$ in infoNCE adopts 1 and shuffling BN is utilized. Momentum SGD is used as the optimizer with an initial learning rate 0.03. The learning rate annealed down to 1e-5 following a cosine decay. The network was trained for 300 epochs with randomly initialized parameters. For data augmentation, we follow \cite{wang2021removing, han2020self} and utilize multi-scale random cropping, random gaussian blur, random color jittering, random temporal jittering \cite{alwassel2019self}, random rotation jittering (randomly rotate 0¡ã-10¡ã), random playback rates and random rotation (random rotate $0\degree,90\degree,180\degree,270\degree$). The size of input video clips is $112\times112$ or $224\times224$, which is noted in the experiments. When jointly optimizing $\mathcal{L}_{ctr}$, $\mathcal{L}_{pcls}$, $\mathcal{L}_{rcls}$ and $\mathcal{L}_{ocls}$, the weights hyper-parameter $\lambda_{ctr}$ is set to 0.1 while other three hyper-parameters $\lambda_{pcls}$, $\lambda_{rcls}$, $\lambda_{ocls}$ are set to 1, based on experimental results.

\textbf{Supervised Fine-tuning Stage.}
In the supervised fine-tuning stage (including linear evaluation and end-to-end fully fine-tuning), for the action recognition task, weights of convolutional layers are retained from the self-supervised model, while weights of fully-connected layers (classifier) are randomly initialized. The network was trained for 100 epochs with cross-entropy loss with supervised labels. The batch size in our implementations was 60 for UCF-101 dataset. The initial learning rate is 0.02 and decay 0.1 when the loss plateaus. The weight decay is same as that in pre-training stage, i.e., 5e-4.

In the inference stage, following common practice in \cite{dave2021tclr}, the final result of a video is the average of the scores of 10 clips that are uniformly sampled from the video.

\textbf{Video Retrieval Settings.}
In the video retrieval stage, we follow the same evaluation protocol described in \cite{luo2020video, xu2019self}. Ten 16-frame clips are uniformly sampled in time from each video and then generate features through a feed-forward pass of a self-supervised pre-trained model with L2 norm. Then, we perform average-pooling over 10 clips to obtain a video-level feature vector. To align the experimental results with prior works \cite{luo2020video, xu2019self} for fair comparison, we use pre-trained models trained based on the proposed self-supervised approach, where no further fine-tuning is allowed. Features in testing splits are regarded as queries and features in training splits are regarded as the gallery. For each feature (query) in the testing splits, Top-k nearest neighbors are recalled from the gallery of training splits by calculating the cosine distances between every two feature vectors.

\subsection{Evaluation on Action Recognition Task}
\subsubsection{End-to-end fine-tuning stage}
Following previous work \cite{chen2021rspnet, jenni2020video}, we compare our proposed method with recent state-of-the-art self-supervised video representation learning approaches. The full results are demonstrated in Table \ref{tb1}, for fair comparison, we report the Top-1 accuracy of UCF-101 and HMDB-51 datasets with commonly used network backbones (C3D, R(2+1)D, S3D) and pre-training datasets, where the results are averaged over three splits of the two datasets. When trained from pre-trained model on UCF-101, it can be observed that our proposed method can achieve state-of-the-art performance among all the approaches with all the backbones. Specifically, our CSTP achieves 85.7\%/55.3\% with R(2+1)D backbone, which outperforms state-of-the-art performance by 4.1\%/6.1\% in UCF-101/HMDB-51 datasets. It even outperforms some approaches trained on Kinetic-400. Similar trend can be observed on C3D and S3D. This demonstrates that our proposed method is generalized to multiple network architectures, which has a robust spatio-temporal modeling ability. When trained from pre-trained model on Kinetics-400, we observe that our CSTP could achieve the state-of-the-art performance in all the backbones. Notably, CSTP unsupervised pre-training leads the accuracy by 0.6\%/1.6\% against fully-supervised ImageNet pre-training with only pre-trained on the small dataset UCF-101 and leads the accuracy by 2.5\%/2.7\% with pre-trained model on Kinetics-400.

When trained from pre-trained model on UCF-101, it can be observed that our proposed method can achieve state-of-the-art performance among all the approaches with all the backbones. Specifically, our CSTP achieves 85.7\%/55.3\% with R(2+1)D backbone, which outperforms state-of-the-art performance by 4.1\%/6.1\% in UCF-101/HMDB-51 datasets. It even outperforms some approaches trained on Kinetic-400. Our CSTP achieved 80.8\%/50.5\% with C3D, which outperformed recent state-of-the-art by 8.4\%/8.2\% in UCF-101/HMDB-51 datasets. And it even outperformed state-of-the-art models that had larger inputs and were trained from Kinetics-400. Similar trend can be observed on S3D. This demonstrates that our proposed method is generalized to multiple network architectures, which has a robust spatio-temporal modeling ability. When trained from pre-trained model on Kinetics-400, we observe that our CSTP could also achieve the state-of-the-art performance in all the backbones. Notably, CSTP unsupervised pre-training leads the accuracy by 0.6\%/1.6\% against fully-supervised ImageNet pre-training with only pre-trained on the small dataset UCF-101 and leads the accuracy by 2.5\%/2.7\% with pre-trained model on Kinetics-400. Using our proposed method, it even outperforms supervised ImageNet pre-training with only pre-trained on small dataset UCF-101 dataset.

\subsubsection{Linear evaluation stage}
Linear evaluation is another straightforward way to qualify the quality of the representation performance of pre-trained models. We also follow the settings of previous work \cite{jenni2020video} for the linear evaluation in which we freeze the convolutional layers and only train the last fully-connected (FC) layers for classification. The fully results of linear evaluation are presented in Table \ref{tb1}, where the ¡°Freeze¡± column in Table \ref{tb1} denotes the linear evaluation setting. We can observe that our method obtains state-of-the-arts performance on UCF-101 and HMDB-51 with multiple backbones (S3D, R3D, C3D) in linear evalution. Specifically, our approach also outperforms TaCo \cite{bai2020can} which combined four existing pretexts tasks with MoCo, demonstrating the representation ability of our proposed method.
\begin{table*}[!t]
	\renewcommand{\arraystretch}{1.5}
	\caption{Comparison with state-of-the-art approaches. Results are the averaged on Top-1 accuracy on \textbf{three splits} in UCF-101 and HMDB-51 datasets.}
	\label{tb1}
	\centering
    \resizebox{17cm}{!}{
    		\begin{tabular}{ccccccccc}
    			\hline\hline \\[-5.5mm]
                \multirow{2}{*}{Method} & \multirow{2}{*}{Year} & \multirow{2}{*}{Backbone} & \multirow{2}{*}{Dataset(duration)} & \multirow{2}{*}{Res.} & \ \multirow{2}{*}{Frames} & \multirow{2}{*}{Freeze} & \multicolumn{2}{c}{Datasets (\%)} \\
                 & & & & & & & UCF-101 & HMDB-51 \\
                \hline
                CBT \cite{sun2019learning} & 2019 & S3D & K600 & 224 & 30 & \textcolor{green}{\Checkmark} & 54.0 & 29.5 \\
                CCL \cite{kong2020cycle} & 2020 & R3D-18 & K400 & 112 & 8 & \textcolor{green}{\Checkmark} & 52.1 & 27.8 \\
                MemDPC \cite{han2020memory} & ECCV 2020 & R3D-34 & K400 & 224 & 40 & \textcolor{green}{\Checkmark} & 54.1 & 27.8 \\
                TaCo \cite{bai2020can} & 2020 & R3D & K400 & 112 & 16 & \textcolor{green}{\Checkmark} & 59.6 & 26.7 \\
                RTT \cite{jenni2020video} & ECCV 2020 & C3D & UCF-101 & 112 & 16 & \textcolor{green}{\Checkmark} & 60.6 & - \\
                MFO \cite{qian2021enhancing} & ICCV 2021 & S3D & K400 & 112 & 16 & \textcolor{green}{\Checkmark} & 61.1 & 31.7 \\
                MFO \cite{qian2021enhancing} & ICCV 2021 & R3D-18 & K400 & 112 & 16 & \textcolor{green}{\Checkmark} & 63.2 & 33.4 \\
                TCLR \cite{dave2021tclr} & 2021 & R3D-18 & UCF-101 & 112 & 16 & \textcolor{green}{\Checkmark} & 67.7 & - \\
                Ours & 2021 & C3D & UCF-101 & 112 & 16 & \textcolor{green}{\Checkmark} & 69.4 & 32.6 \\
                Ours & 2021 & R3D & K400 & 112 & 16 & \textcolor{green}{\Checkmark} & 70.5 & 34.4 \\
                Ours & 2021 & S3D & K400 & 112 & 16 & \textcolor{green}{\Checkmark} & 71.4 & 35.3 \\
                Ours & 2021 & R(2+1)D & K400 & 112 & 16 & \textcolor{green}{\Checkmark} & \textbf{75.3} & \textbf{39.2} \\
                \hline
                \multicolumn{2}{c}{Random Initialization (scratch)} & R(2+1)D & - & 112 & 16 & \textcolor{red}{\XSolidBrush} & 60.3 & 27.6 \\
                \multicolumn{2}{c}{Supervised ImageNet Pre-training} & R(2+1)D & ImageNet & 112 & 16 & \textcolor{red}{\XSolidBrush} & 85.1 & 53.7 \\
                \hline
                VCP \cite{luo2020video} & AAAI 2020 & R(2+1)D & UCF-101 (1d) & 112 & 16 & \textcolor{red}{\XSolidBrush} & 66.3 & 32.2 \\
                PRP \cite{yao2020video} & CVPR 2020 & R(2+1)D & UCF-101 (1d) & 112 & 16 & \textcolor{red}{\XSolidBrush} & 72.1 & 35.0 \\
                VCOP \cite{xu2019self} & CVPR 2019 & R(2+1)D & UCF-101 (1d) & 112 & 16 & \textcolor{red}{\XSolidBrush} & 72.4 & 30.9 \\
                Var.PSP \cite{cho2020self} & 2020 & R(2+1)D & UCF-101 (1d) & 112 & 16 & \textcolor{red}{\XSolidBrush} & 74.8 & 36.8 \\
                PacePred \cite{wang2020self} & ECCV 2020 & R(2+1)D & UCF-101 (1d) & 112 & 16 & \textcolor{red}{\XSolidBrush} & 75.9 & 35.9 \\
                RSPNet \cite{chen2021rspnet} & ECCV 2020 & R(2+1)D & K400 (28d) & 112 & 16 & \textcolor{red}{\XSolidBrush} & 81.1 & 44.6 \\
                RTT \cite{jenni2020video} & ECCV 2020 & R(2+1)D & UCF-101 (1d) & 112 & 16 & \textcolor{red}{\XSolidBrush} & 81.6 & 46.4 \\
                VideoMoCo \cite{pan2021videomoco} & CVPR 2021 & R(2+1)D & K400 (28d) & 112 & 16 & \textcolor{red}{\XSolidBrush} & 78.7 & 49.2 \\
                Ours CSTP & - & R(2+1)D & UCF-101 (1d) & 112 & 16 & \textcolor{red}{\XSolidBrush} & 85.7 & 55.3 \\
                Ours CSTP & - & R(2+1)D & K400 (28d) & 112 & 16 & \textcolor{red}{\XSolidBrush} & \textbf{87.6} & \textbf{56.4} \\
                \hline
                RTT \cite{jenni2020video} & ECCV 2020 & C3D & UCF-101 (1d) & 112 & 16 & \textcolor{red}{\XSolidBrush} & 68.3 & 38.4 \\
                PRP \cite{yao2020video} & CVPR 2020 & C3D & UCF-101 (1d) & 112 & 16 & \textcolor{red}{\XSolidBrush} & 69.1 & 34.5 \\
                VCP \cite{luo2020video} & AAAI 2020 & C3D & UCF-101 (1d) & 112 & 16 & \textcolor{red}{\XSolidBrush} & 68.5 & 32.5 \\
                VCOP \cite{xu2019self} & CVPR 2019 & C3D & UCF-101 (1d) & 112 & 16 & \textcolor{red}{\XSolidBrush} & 65.6 & 28.4 \\
                Var.PSP \cite{cho2020self} & 2020 & C3D & UCF-101 (1d) & 112 & 16 & \textcolor{red}{\XSolidBrush} & 70.4 & 34.3 \\
                MoCo+BE \cite{wang2021removing} & CVPR 2021 & C3D & UCF-101 (1d) & 112 & 16 & \textcolor{red}{\XSolidBrush} & 72.4 & 42.3 \\
                RSPNet \cite{chen2021rspnet} & AAAI 2021 & C3D & K400 (28d) & 112 & 16 & \textcolor{red}{\XSolidBrush} & 76.7 & 44.6 \\
                Ours CSTP & - & C3D & UCF-101 (1d) & 112 & 16 & \textcolor{red}{\XSolidBrush} & 80.8 & 48.9 \\
                Ours CSTP & - & C3D & K400 (28d) & 112 & 16 & \textcolor{red}{\XSolidBrush} & \textbf{82.6} & \textbf{50.2} \\
                \hline
                MFO \cite{qian2021enhancing} & ICCV 2021 & S3D & UCF-101 (1d) & 112 & 16 & \textcolor{red}{\XSolidBrush} & 74.3 & 37.2 \\
                CoCLR \cite{han2020self} & NIPS 2020 & S3D & UCF-101 (1d) & 112 & 16 & \textcolor{red}{\XSolidBrush} & 81.4 & 52.1 \\
                Ours CSTP & - & S3D & UCF-101 (1d) & 112 & 16 & \textcolor{red}{\XSolidBrush} & 83.1 & 51.9 \\
                Ours CSTP & - & S3D & UCF-101 (1d) & 128 & 16 & \textcolor{red}{\XSolidBrush} & \textbf{83.6} & \textbf{53.3} \\
                \hline
    			\hline\hline
    		\end{tabular}
     }
\end{table*}

\subsection{Evaluation on video retrieval}
Besides the video action recognition task, we also report the Top-k (k=1, 5, 10, 20, 50) video retrieval performance with two different backbones, i.e., R(2+1)D and C3D. Our CSTP is pre-trained on UCF-101 dataset. The full quantitative results are shown in Table \ref{tb2}. We observe that our proposed approach outperforms state-of-the-art approaches by a large margin under different k with the two network backbones. Specifically, our CSTP outperforms the state-of-the-art model by 16.6\%/6.9\% in UCF-101/HMDB-51 dataset using R(2+1)D as the backbone, and outperforms the state-of-the-art model by 14.7\%/6.7\% in UCF-101/HMDB-51 dataset using C3D as the backbone. These imply that our proposed CSTP could help the model to learn more discriminating spatio-temporal features for the video retrieval task.

\begin{table*}[!t]
	\renewcommand{\arraystretch}{1.5}
	\caption{Recall-at-Topk. Comparison with state-of-the-art methods in video retrieval task on UCF-101 and HMDB-51.}
	\label{tb2}
	\centering
    \resizebox{17cm}{!}{
    		\begin{tabular}{ccccccccccccc}
    			\hline\hline \\[-5.5mm]
                \multirow{2}{*}{Method} & \multirow{2}{*}{Dataset} & \multirow{2}{*}{Backbone} & \multicolumn{5}{c}{UCF-101} & \multicolumn{5}{c}{HMDB-51} \\
                 & & & R@1 & R@5 & R@10 & R@20 & R@50 & R@1 & R@5 & R@10 & R@20 & R@50 \\
                 SpeedNet \cite{benaim2020speednet} & K400 & S3D-G & 13.0 & 28.1 & 37.5 & 49.5 & 65.0 & - & - & - & - & - \\
                 RSPNet \cite{chen2021rspnet} & UCF-101 & R3D-18 & 41.1 & 59.4 & 68.4 & 77.8 & 88.7 & - & - & - & - & - \\
                 MFO \cite{qian2021enhancing} &  UCF-101 & R3D-18 & 39.6 & 57.6 & 69.2 & 78.0 & - & 18.8 & 39.2 & 51.0 & 63.7 & - \\
                 \hline
                 VCOP \cite{xu2019self} & UCF-101 & R(2+1)D & 10.7 & 25.9 & 35.4 & 47.3 & 63.9 & 5.7 & 19.5 & 30.7 & 45.6 & 67.0 \\
                 VCP \cite{luo2020video} & UCF-101 & R(2+1)D & 19.9 & 33.7 & 42.0 & 50.5 & 64.4 & 6.7 & 21.3 & 32.7 & 49.2 & 73.3 \\
                 PRP \cite{yao2020video} & UCF-101 & R(2+1)D & 20.3 & 34.0 & 41.9 & 51.7 & 64.2 & 8.2 & 25.3 & 36.2 & 51.0 & 73.0 \\
                 PacePred \cite{wang2020self} & UCF-101 & R(2+1)D & 25.6 & 42.7 & 51.3 & 61.3 & 74.0 & 12.9 & 31.6 & 43.2 & 58.0 & 77.1 \\
                 IIC \cite{tao2020self} & UCF-101 & R(2+1)D & 34.7 & 51.7 & 60.9 & 69.4 & 81.9 & 12.7 & 33.3 & 45.8 & 61.6 & 81.3 \\
                 Our CSTP & UCF-101 & R(2+1)D & \textbf{51.3} & \textbf{71.1} & \textbf{80.3} & \textbf{86.8} & \textbf{93.8} & \textbf{21.6} & \textbf{48.4} & \textbf{62.4} & \textbf{74.7} & \textbf{89.4} \\
                 \hline
                 VCOP \cite{xu2019self} & UCF-101 & C3D & 12.5 & 29.0 & 39.0 & 50.6 & 66.9 & 7.4 & 22.6 & 34.4 & 48.5 & 70.1 \\
                 VCP \cite{luo2020video} & UCF-101 & C3D & 17.3 & 31.5 & 42.0 & 52.6 & 67.7 & 7.8 & 23.8 & 35.3 & 49.3 & 71.6 \\
                 PRP \cite{yao2020video} & UCF-101 & C3D & 23.2 & 38.1 & 46.0 & 55.7 & 68.4 & 10.5 & 27.2 & 40.4 & 56.2 & 75.9 \\
                 PacePred \cite{wang2020self} & UCF-101 & C3D & 31.9 & 49.7 & 59.2 & 68.9 & 80.2 & 12.5 & 32.2 & 45.4 & 61.0 & 80.7 \\
                 IIC \cite{tao2020self} & UCF-101 & C3D & 31.9 & 48.2 & 57.3 & 67.1 & 79.1 & 11.5 & 31.3 & 43.9 & 60.1 & 80.3 \\
                 RSPNet \cite{chen2021rspnet} & UCF-101 & C3D & 36.0 & 56.7 & 66.5 & 76.3 & 87.7 & - & - & - & - & - \\
                 MoCo+BE \cite{wang2021removing} & UCF-101 & C3D & - & - & - & - & - & 10.2 & 27.6 & 40.5 & 56.2 & 76.6 \\
                 Our CSTP & UCF-101 & C3D & \textbf{50.7} & \textbf{69.4} & \textbf{77.9} & \textbf{84.6} & \textbf{91.1} & \textbf{19.3} & \textbf{46.6} & \textbf{61.0} & \textbf{71.2} & \textbf{86.3} \\
    			\hline\hline
    		\end{tabular}
     }
\end{table*}

\end{document}